\documentclass[11pt,a4paper]{article}
\usepackage{times,latexsym}
\usepackage{url}
\usepackage[T1]{fontenc}

\usepackage{graphicx}
\usepackage{amsmath}
\usepackage{multirow}

\usepackage[acceptedWithA]{tacl2018v2}
\usepackage[]{tacl2018v2}

\usepackage{xspace,mfirstuc,tabulary}

\newif\iftaclinstructions
\taclinstructionsfalse 
\iftaclinstructions
\renewcommand{\confidential}{}
\renewcommand{\anonsubtext}{(No author info supplied here, for consistency with
TACL-submission anonymization requirements)}
\newcommand{\instr}
\fi

\iftaclpubformat 

\else

\fi

\title{Learning End-to-End Goal-Oriented Dialog with Maximal User Task Success and Minimal Human Agent Use}

\author{
  Janarthanan Rajendran\thanks{\hspace{0.2cm}This work was done when the author was an intern at IBM Research, NY.}\\
  University of Michigan \\
  {\tt rjana@umich.edu}
  \\\And
  Jatin Ganhotra \\
  IBM Research \\
  {\tt jatinganhotra@us.ibm.com}
  \\\And
  Lazaros Polymenakos\thanks{\hspace{0.2cm}This work was done when the author was at IBM Research, NY.}\\
  Amazon Alexa-AI Research\\
  {\tt polyml@amazon.com}
  }

\date{}

\begin{document}
\maketitle
\begin{abstract}
  Neural end-to-end goal-oriented dialog systems showed promise to reduce the workload of human agents for customer service, as well as reduce wait time for users. However, their inability to handle new user behavior at deployment has limited their usage in real world. In this work, we propose an end-to-end trainable method for neural goal-oriented dialog systems which handles new user behaviors at deployment by transferring the dialog to a human agent intelligently. The proposed method has three goals: 1) maximize user's task success by transferring to human agents, 2) minimize the load on the human agents by transferring to them only when it is essential and 3) learn online from the human agent's responses to reduce human agents load further. We evaluate our proposed method on a modified-bAbI dialog task that simulates the scenario of new user behaviors occurring at test time. Experimental results show that our proposed method is effective in achieving the desired goals.
\end{abstract}

\iftaclpubformat

\section{Introduction}
Neural end-to-end dialog systems showed huge potential for various goal-oriented dialog tasks such as restaurant reservation, flight ticket booking and hotel reservation. However, their use in the real world has been limited due to the inability to handle new user behavior at deployment.

There are two main methods to build neural end-to-end goal-oriented dialog systems. In the first method, large amounts of human-human chat logs of a particular task are collected and then the dialog system is trained to mimic the chat logs using Supervised Learning (SL) (\citet{bordes2016learning}). In the second method, the dialog system is trained to complete the task against a human (user) simulator ( \citet{young2013pomdp}). The training is done using Reinforcement Learning (RL) by providing reward for task completion and also intermediate rewards for pre-identified sub-task completion. This is often accompanied by a SL pre-training as in \citet{lui_lane_2017}.

\begin{figure*}[ht]
\centering
\includegraphics[width=1\textwidth]{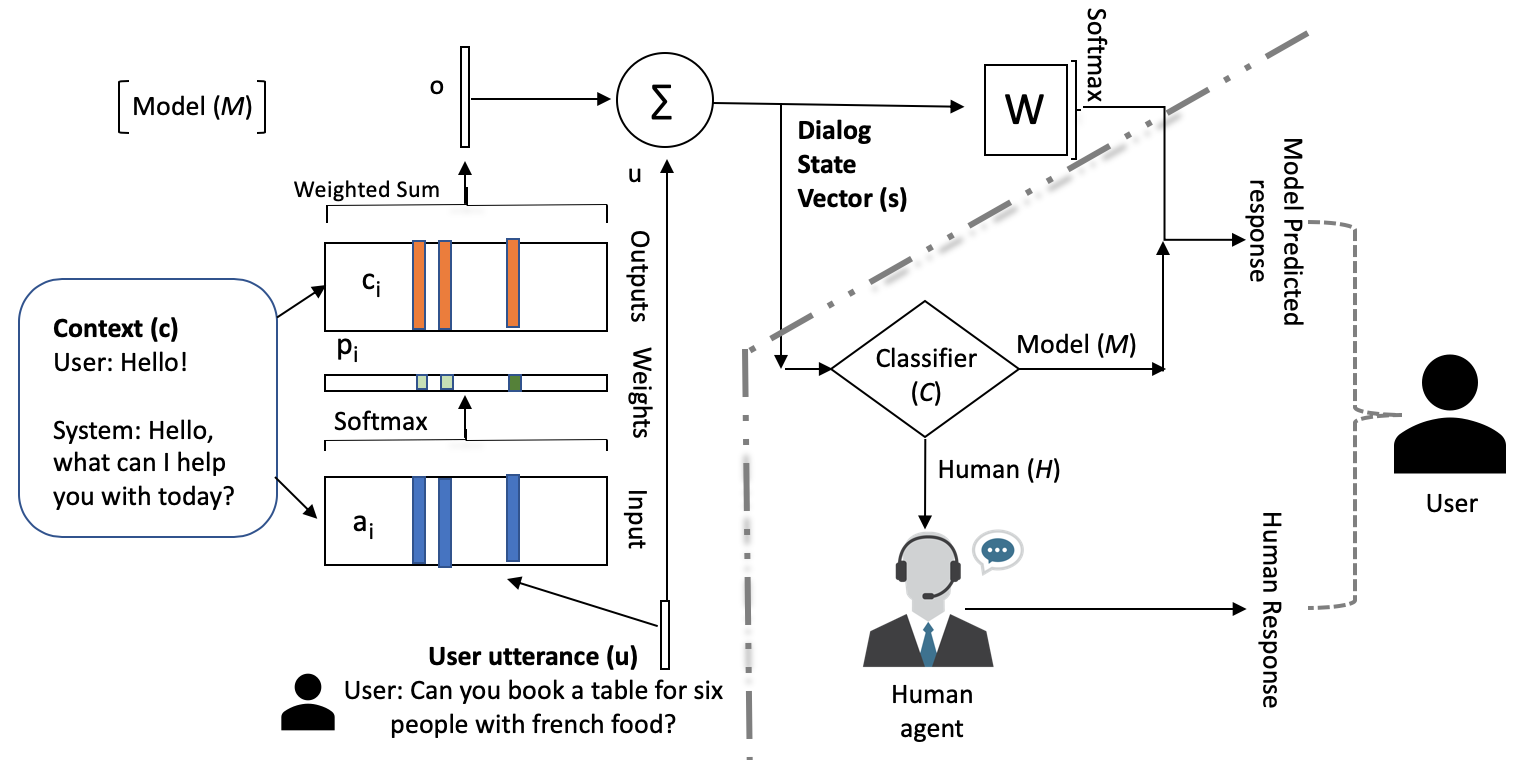}
\caption{\textbf{\textit{Left}}: A single layer version of memN2N (model $M$) \textbf{\textit{Right}}: Proposed Method} 
\label{fig_model}
\end{figure*}

Dialog systems built using either of these methods would fail in the presence of new user behaviors during deployment, which were missing in training. Failure here refers to the inability of the dialog system to complete the task for the user. The new behaviors can be a different way of a user asking/providing certain information or could also be as simple as an user utterance with Out-Of-Vocabulary (OOV) words. The failure happens when these new behaviors are beyond the generalization capabilities of the trained systems due to the limited coverage of training data collected. For a real-world use-case, it is difficult to collect chat logs and/or to build a user simulator that covers all possible user behaviors, which implies that users with new behaviors are bound to come by during deployment. The new user behaviors that we refer to and focus here are only those that are valid and acceptable, i.e. these user behaviors should ideally have been part of the training data (chat logs in the case of SL and user simulator behavior pattern in the case of RL).

For an enterprise that uses this dialog system, these failures could affect their business directly. In addition to losing customers who faced these failures, the enterprise also loses future users/customers, as it also affects the perceived reliability of the dialog system and hence, the enterprise's reputation. While the dialog system fails for new user behaviors, it can perform the task well for majority of user behaviors. However, these failures have restricted the deployment of neural end-to-end goal-oriented dialog systems and have forced the enterprises to rely completely on human agents or other systems.

There have been recent works on using user feedback (\citet{lui_lane_2017}) and active human teaching (\citet{hitlDL}) during deployment to improve robustness to new user behaviors. While they have produced better systems, they are not good enough to be deployed in the real world. In this work, we propose a method which can be used in addition to the aforementioned ideas and enables a dialog system to perform well in the real world.

We propose an end-to-end trainable method, in which the dialog system can automatically identify a new user behavior during deployment that the system might fail at and transfer the task to a human agent, such that the user's task is completed without any issue. At the same time, the dialog system also learns from the human agent's response to handle that new user behavior in future. Our method also allows one to choose the trade-off between maximizing user's task success and minimizing the workload on human agents.

We set the following three goals for our method:
\begin{itemize}
    \item Maximize task success rate for the user by transferring to a human agent in cases where the dialog system might fail
    \item Minimize the use of human agent by transferring to the human agent only when it is essential
    \item Learn online from the human agent's response to reduce the use of human agent over time
\end{itemize}

This paper is organized as follows. Section \ref{proposed_method} describes the proposed method. In Section \ref{modified_babi}, we introduce modified-bAbI dialog tasks, which simulate new user behaviours occurrence at deployment and serve as testbed for evaluating our proposed method. Section \ref{experiments_and_results} details our experimental results on modified-bAbI dialog tasks. Section \ref{related_work} discusses related work and Section \ref{conclusion} concludes.

\section{Proposed Method}
\label{proposed_method}
Consider a neural dialog model $M$ trained for a goal-oriented dialog task. We also have a human agent $H$ who is trained for the same task. Both $M$ and $H$ can take the dialog so far as input and produce the response for the user utterance ($u$). There is a neural classifier $C$ which uses the dialog state vector ($s$) from model $M$ as input and decides whether to use the model $M$ to provide response to the user or to transfer to the human agent $H$ who could then provide the response to the user. The proposed method is shown in Fig \ref{fig_model} (right).

In a real world setting, we cannot expect the same exact user utterances to come during deployment, that the model came across during its training. Therefore, for a new dialog, it is not possible to know beforehand if model ($M$) would provide a correct response or not. The classifier ($C$) has to learn this through trial and error and generalize. Therefore, the classifier is trained using RL.

The classifier is provided a high reward if it chooses the model $M$ and the model produces a correct/valid response. The classifier is however provided a relatively smaller reward if it chooses the human agent instead. We assume that the human agent's response is always correct. If the classifier chooses the model $M$ and the model provides an incorrect response, the classifier ($C$) is penalized heavily. The validation, if a response is correct or not is provided by the user as feedback. The classifier is trained using RL to make decisions (take actions) in order to maximize the above reward function. The reward function helps achieve two of our aforementioned goals -
\begin{itemize}
    \item \textit{Maximize task success rate for the user}: The reward function encourages the classifier to learn the dialog scenarios in which the model $M$ might fail and choose a human agent instead. Therefore, the classifier helps to avoid sending an incorrect response to the user from the model.
    \item \textit{Minimize human agent use}: The reward function also encourages the classifier to learn, identify and choose the model $M$ in cases where the model has a high chance of providing the correct response, as the classifier gets a higher reward compared to choosing a human agent. This minimizes the use of human agent to only when it is essential. 
\end{itemize}

Here is an example reward function, which would achieve the desired goals:
\begin{itemize}
    \item +1 : if human $H$ is chosen
    \item +2 : if model $M$ is chosen and the model's response is correct
    \item -4 : if model $M$ is chosen and the model's response is incorrect
\end{itemize}

The reward function allows the designer to choose the trade off between maximizing the user task completion vs minimizing the human agent's workload. For example, when the model ($M$) is chosen, increasing the positive reward if the model's response is correct and reducing the penalty when the model's response is incorrect would encourage the overall system to use model ($M$) more to respond to the user.

The gradient updates obtained for the classifier through the rewards received are also propagated back to the model $M$ through the dialog state vector. This trains the model $M$ to incorporate a) essential information about the dialog so far and b) the model's confidence in producing the correct response, in the dialog state vector, such that the classifier can utilize it to make the right decision. 

Whenever the classifier chooses the human agent ($H$), the dialog interaction (including the human response) is also added to the training data of the model ($M$) and the model is updated online using supervised learning (SL). This helps achieve our third goal:
\begin{itemize}
    \item \textit{Reduce human agent use over time}: The online update allows the model $M$ to respond to the corresponding new user behavior and provide the correct response if a similar dialog scenario occurs in the future. This also enables the classifier to reduce its dependence on the human agent ($H$) over time. 
\end{itemize}
 
The classifier keeps changing during its lifetime to adapt to the changes in the model $M$. Note that a human agent is involved only when the classifier transfers the dialog to a human. The idea is generic enough to be used with any neural dialog model ($M$), e.g. HRED \cite{sordoni2015hierarchical}, end-to-end memory network \cite{sukhbaatar2015end} etc. In our experiments, we use the end-to-end memory network for our model $M$ and a Multi-Layer Perceptron (MLP) for the classifier $C$. Additional details on the model architecture and training are provided in Section \ref{experiments_and_results}.


\begin{figure*}[ht]
\centering
\includegraphics[scale=0.5]{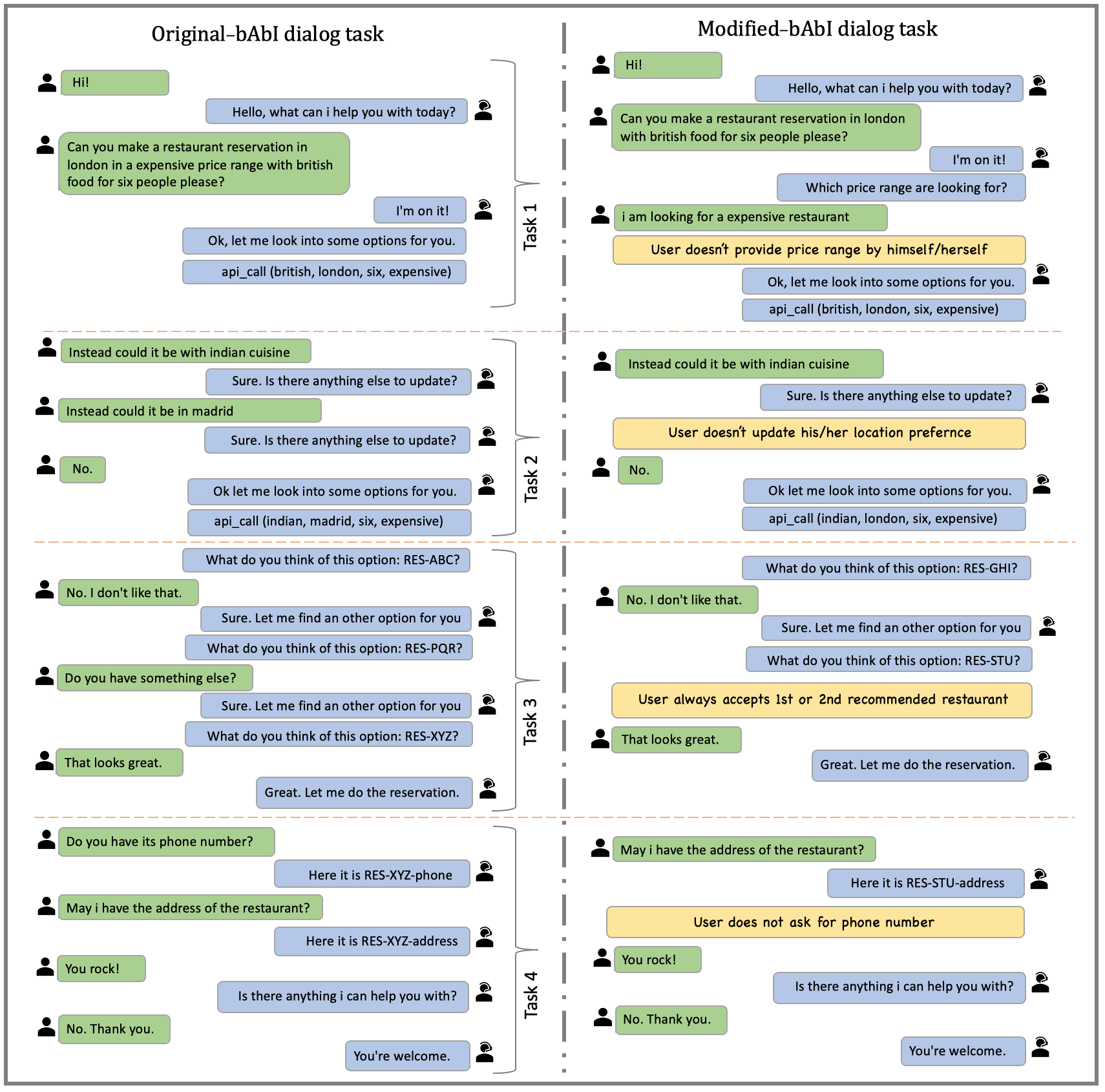}
\caption{\textbf{Modified-bAbI dialog task}. A user (in green) chats with a dialog system (in blue) to book a table at a restaurant. 
We update each subtask in bAbI dialog task with specific changes (in yellow).}
\label{fig_dataset}
\end{figure*}


\section{Modified bAbI dialog tasks}
\label{modified_babi}
bAbI dialog tasks (referred to as original-bAbI dialog tasks from here on) were proposed by \cite{bordes2016learning} as a testbed to evaluate the strengths and shortcomings of end-to-end dialog systems in goal-oriented applications (\citet{seo2016query}, \citet{rajendran2018learning}). The dataset is generated by a restaurant reservation simulation where the final goal is booking a table. The simulator uses a Knowledge Base (KB) which contains information about restaurants. There are five tasks: Task 1 (Issuing API calls), Task 2 (Updating API calls), Task 3 (Displaying options), Task 4 (Providing extra information) and Task 5 (Conducting full dialogs). Tasks 1 and 2 test the dialog system to implicitly track dialog state, Tasks 3 and 4 check if the system can learn to use the information from KB and Task 5 combines all tasks. It is a retrieval task, where the dialog system has to retrieve the correct response from a set of given candidate responses.

We propose modified-bAbI dialog tasks, an extension of original-bAbI dialog tasks \citep{bordes2016learning}. We modify the original-bAbI dialog tasks, by removing and replacing certain user behaviors from the training and validation data. The test set is left untouched. This simulates a scenario where some new user behaviors arise during the test (deployment) time that were not seen during the training and hence allows us to test our proposed method. This also mimics real-world data collection via crowdsourcing in the sense that certain user behavior is missing from the training data. Fig \ref{fig_dataset} shows a dialog sample from modified-bAbI dialog tasks.\footnote{modified-bAbI dialog tasks - \url{https://github.com/IBM/modified-bAbI-dialog-tasks}} We propose the following changes to the training and validation data of the original-bAbI dialog tasks:

\begin{table}[ht]
\begin{center}
\begin{tabular}{c|c}
    \hline
    \textbf{Behavior type} & \textbf{ \# dialogs}\\ \hline
    Task-1 & 494\\ \hline
    Task-2 & 539\\ \hline
    Task-3 & 561\\ \hline
    Task-4 & 752\\ \hline
\end{tabular}
\end{center}
\caption{Modified-bAbI test set statistics. The numbers shown represent number of dialogs in test data (Task 5) out of 1000 dialogs where a certain type of new user behavior is encountered.}
\label{tab:tst-dataset-stats-1}
\end{table}

\begin{table}[h]
\begin{center}
\begin{tabular}{c|c|c|c|c|c}
    \hline
    \textbf{\# new behavior} & \textbf{0} & \textbf{1} & \textbf{2} & \textbf{3} & \textbf{4}\\ \hline
    \textbf{\# dialogs} & 20 & 178 & 350 & 340 & 112\\ \hline
\end{tabular}
\end{center}
\caption{Modified-bAbI test set statistics. The numbers shown represent number of dialogs in test data (Task 5) out of 1000 dialogs, where no new user behavior or one or more type of new user behavior is encountered.}
\label{tab:tst-dataset-stats-extended}
\end{table}


\begin{itemize}
    \item In Task 1, a user places a request that can contain from 0 to 4 of the required fields to book a table. The system asks questions to retrieve values for the missing fields and generate the correct corresponding API call. In modified-bAbI dialog tasks, the user doesn't provide the value for the price range by himself/herself and only provides that information when asked by the system.
    \item In Task 2, the user can request the system to update any of his/her preferences (cuisine, location, price range and number of people). In modified-bAbI dialog tasks, the user doesn't update his/her location preference.
    \item In Task 3, for the API call matching the user request, information retrieved from the KB is provided as part of dialog history. The system must propose options to the user by listing the restaurant names sorted by their corresponding rating (in decreasing order). The system keeps proposing a new restaurant until the user accepts. In modified-bAbI dialog tasks, the user always accepts first or second recommended restaurant.
    \item In Task 4, the user can ask for phone number or address for the selected restaurant. In modified-bAbI dialog tasks, the user does not ask for phone number.
\end{itemize}

We incorporate the changes mentioned above to the final Task 5 (Conducting full dialogs). We perform experiments on modified-bAbI dialog task-5, as tasks 1-4 are subsets of a full conversation and don't represent a complete meaningful real-world conversation standalone. The statistics for new user behavior in the test set (which was left untouched) are shown in Table \ref{tab:tst-dataset-stats-1} and \ref{tab:tst-dataset-stats-extended}.

\section{Experiments and Results}
\label{experiments_and_results}
\subsection{Baseline method:}
\label{baseline-test-results}
A dialog model $M$ is trained on the modified dialog bAbI task and is used for deployment. The model is not updated during test/deployment. An end-to-end memory network \cite{sukhbaatar2015end} architecture is used for the model $M$. End-to-end memory networks (memN2N) are an extension of Memory Networks proposed by \cite{weston2014memory}. They have been successful on various natural language processing tasks and perform well on original-bAbI dialog tasks. Hence, we chose them for our model $M$.

A single layer version of the memN2N model is shown in Fig.\ref{fig_model} (left). A given sentence $(i)$ from the context (dialog history) is stored in the memory by: a) its input representation $(a_{i})$ and b) its output representation $(c_{i})$, where each memory contains the embedding representation for that sentence. The embedding representation of the sentence is calculated by adding the embeddings of all the words in the sentence \cite{bordes2016learning}. Attention of current user utterance (u) over memory is computed via dot product, to identify the relevance of the memory w.r.t the current user utterance (query). $(p_{i})$ represents the probability for each memory based on attention scores (equation \ref{eq3}). An output vector $(o)$ is computed by the weighted sum of memory embeddings $(c_{i})$ with their corresponding probabilities $(p_{i})$ (equation \ref{eq4}). The output vector $(o)$ is the overall context embedding and, with query $(u)$ represents the dialog state vector $(s)$ (equation \ref{eq5}). The attention can be done multiple times, i.e. multiple layers (3 in our experiments) by updating $u$ with $s$ and repeating equations \ref{eq3}, \ref{eq4} and \ref{eq5}. The last internal state is then used as the dialog state vector $s$ to select the candidate response for the model $M$ and also provided as input to the classifier $C$.
\begin{align}
p_{i} &= \textrm{Softmax}(u^{T} (a_{i})) \label{eq3} \\
o &= \sum_{i}p_{i}c_{i} \label{eq4} \\
s &= (o + u) \label{eq5}
\end{align}

The model is trained using SL on the training data. The trained model is then used during deployment. In our case, deployment is same as testing the model on the test data. Our results for the baseline method for the original and modified bAbI dialog tasks are given in Table \ref{tab:baseline-results}. The hyper-parameters used for training the model $M$ in the baseline method are provided in Appendix \ref{appendix_a1}.

Per-turn accuracy measures the percentage of responses that are correct (i.e., the correct candidate is selected from all possible candidates). Note that, as mentioned above in Section \ref{modified_babi}, we only modify the training and validation sets, and use the same test set. Per-dialog accuracy measures the percentage of dialogs where every response is correct. Therefore, even if only one response in a dialog is incorrect, this would result in a failed dialog, i.e. failure to achieve the goal of booking a table.

\begin{table}[h]
\begin{center}
\begin{tabular}{c|c|c}
    \hline
    \textbf{Dataset} & \textbf{Per-turn} & \textbf{Per-dialog} \\ \hline
    Original-bAbI & 98.5 & 77.1 \\ \hline
    Modified-bAbI & 81.7 & 3.7 \\ \hline
\end{tabular}
\end{center}
\caption{Test results (accuracy \%) for our baseline method ($M$:memN2N) across original and modified bAbI dialog tasks.}
\label{tab:baseline-results}
\end{table}

From Table \ref{tab:baseline-results}, we observe that the baseline method of using the trained memN2N model performs poorly on modified-bAbI dialog tasks, which has new user behaviors at test time that the model has not seen during training. For modified-bAbI dialog tasks, the baseline method achieves 81.7\% on per-turn accuracy and the per-dialog accuracy decreases to 3.7\%. This implies that majority of dialogs would be unsuccessful in completing the goal. These results clearly demonstrate that the baseline method (end-to-end memory network) does not perform well on our proposed testbed, which simulates new user behaviours occurrence at deployment.

\subsection{Proposed Method: $M^*+C^*$}
We use the same memN2N model used in the baseline method for the model $M$ here. However, in our proposed method, we also have a classifier $C$ which takes as input the dialog state vector $s$ and makes a decision on whether to use the model to respond to the user or to transfer the dialog to a human agent. For our experiments, $C$ is a Multi-Layer-Perceptron (MLP) which outputs a probability distribution over the two actions. The sampled action is performed and the user feedback is used to determine the reward obtained, which is then used to train the classifier and the model.

\begin{table*}[htbp!]
    \centering
    \begin{tabular}{ll|l|l|l|l|l}
    \hline
    \multicolumn{2}{c|}{\textbf{Method}} & \multicolumn{2}{c|}{\textbf{User Accuracy}} & \textbf{ Model ratio} & \multicolumn{2}{c}{\textbf{Final Model Accuracy}} \\ 
    & & \textbf{Per-turn} & \textbf{Per-dialog} & & \textbf{Per-turn} & \textbf{Per-dialog} \\\hline
    \multicolumn{2}{c|}{Baseline method (M)} & 81.73(0) & 3.7(0) & 100.0(0) & 81.73(0) & 3.7(0) \\\hline
    \multirow{2}{*}{R: 1, 2, -4} & $M+C^*$ & 92.85(1.58) & 33.48(10.59) & 51.97(8.22) & 81.73(0) & 3.7(0) \\\cline{2-7}
    &$M^*+C^*$ & 96.28(1.16) & 54.5(10.72) & 64.06(4.65) & 90.83(0.82) & 14.82(3.7) \\\cline{2-7}
    &$M_a^*+C^*$ & 96.19 (1.21) & 54.44(11.40) & 61.14(6.9) & 88.98(0.34) & 10.26(1.39) \\\hline
    \hline
    \multirow{2}{*}{R: 1, 3, -3} & $M+C^*$ & 91.31(1.15) & 26.50(7.57) & 58.82(4.62) & 81.73(0) & 3.7(0) \\\cline{2-7}
    & $M^*+C^*$ & 94.67(1.20) & 43.48(8.80) & 70.33(2.13) & 89.27(0.74) & 12.84(2.22) \\\cline{2-7}
    & $M_a^*+C^*$ & 94.08(1.0) & 38.8(8.15) & 69.69(6.14) & 88.75(0.91) & 11.62(2.61) \\\hline
    \end{tabular}
    \caption{\textbf{Test results for the different methods on the modified-bAbI dialog task.} The numbers represent the mean and standard deviation(shown in parenthesis) of running the different methods across 5 different permutations of the test set. \textit{User Accuracy}: Task success rate for the user; \textit{Model ratio}: Percentage of time the classifier chooses the model $M$; \textit{Final Model Accuracy}: Accuracy of the model $M$ at the end of testing. }
    \label{tab:proposed_method}
\end{table*}

The following scenarios arise during deployment depending on the classifier's action and the model's prediction:
\begin{itemize}
    \item The classifier could choose a human agent $H$ to respond to the user. Since we use the test data of the modified-bAbI dialog task as a way to test deployment performance, we already have the ground truth labels for the different dialog scenarios that arise during the deployment. We use these ground truth labels as the human response.
    \item The classifier chooses the model $M$ and the model produces a correct or incorrect response. In real world, this validation/feedback on whether the response was correct or not is obtained from the user during deployment. For our experiments, we use the ground truth labels for test data to provide the validation from the user. In a sense, we mimic an ideal user using the test data ground truth labels.
\end{itemize}

We have two versions of the proposed method: a) the model trained on the training data is kept fixed during deployment and only the classifier is updated ($M+C^*$) b) both the model and the classifier are updated ($M^* + C^*$) during deployment. For both versions, the classifier $C$ is initialized randomly and is updated only during deployment. We use REINFORCE (\citet{williams1992simple}) for training the classifier using the rewards obtained.

For $M^* + C^*$, the model is updated using the following three ways:
\begin{itemize}
    \item The gradients obtained for the classifier $C$ are passed through the dialog state vector $s$ to the model. 
    \item The human responses provided for cases where the classifier transferred the dialog to a human agent, are added to the training data to augment it and are also used to update the model using supervised learning.
    \item Dialogs are sampled from the augmented training data and are used to update the model using supervised learning to avoid forgetting. 
\end{itemize} 

The $M^*+C^*$ method 
uses a fixed number of samples, e.g. 2 batches for our experiments, from augmented training data to update the model. We also implement and evaluate a variant of $M^* + C^*$ method: $M_a^* + C^*$ where the number of samples are decided based on the model $M$'s performance on validation data. During deployment, after each batch of test data, the model is evaluated on the validation data. The difference between the current validation per-turn accuracy ($v^{current}_{acc}$) and the best validation per-turn accuracy so far ($v^{best}_{acc}$) estimates the loss in information learned from original training data during deployment. This is used to determine the number of batches ($b\geq0$) for updating the model, as per the equation:
\begin{align}
\label{eq6}
b &= \alpha * (v^{current}_{acc} - v^{best}_{acc})
\end{align}

The $M_a^* + C^*$ ($a$ - adaptive) method tries to update the model with the right number of dialog samples from the augmented data. Further details on training are provided in Appendix \ref{appendix_a2}. We perform our experiments on two sets of reward functions. The first reward function is as follows:
\begin{itemize}
    \item +1 : if human $H$ is chosen
    \item +2 : if model $M$ is chosen and the model's response is correct
    \item -4 : if model $M$ is chosen and the model's response is incorrect
\end{itemize}

The results are shown in Table \ref{tab:proposed_method}. The test data is provided sequentially, which mimics the deployment in a real-world setting. Since the performance depends on the order in which the test dialogs are provided, we evaluate our proposed method on 5 different permutations of the test set. We present the mean and standard deviation of the performance measures across the 5 permutations. The results for the individual permutations are provided in Appendix \ref{appendix_b}. The performance measures used for evaluation are:
\begin{itemize}
    \item \textit{User Accuracy}: Task success rate as perceived by the user, irrespective of whether the response is provided by the human agent $H$ or the model $M$. This measures our goal \#1 - Maximize task success rate
    \item \textit{Model ratio}: Percentage of time the classifier $C$ provides the model response to the user, i.e human agent is not involved. This measures our goal \#2 - Minimize human agent use
    \item \textit{Final Model Accuracy}: Accuracy of the model $M$ on the test data at the end of testing. This is obtained by evaluating the model $M$ on the test data again after the testing phase is over.  This measures our goal \#3 - Reduce human agent use over time, by online learning of the model.
\end{itemize}

While the baseline method $M$ gets a per-turn user accuracy of 81.73\%, using and learning a classifier ($M + C^*$) achieves user accuracy of 92.85\%, an increase of more than 10 percentage points. If the model is also updated during the deployment ($M^* + C^*$), we observe further increase in per-turn accuracy (96.28\%). While $M+C^*$ achieves better performance by using the model 51.97\% of the time, $M^*+C^*$ achieves high accuracy by using model even more (64.06\%), thereby reduces the human agent's workload.  This is attributed to the improvement in the model during the deployment for the $M^*+C^*$ method. This is supported by the improvement in the model accuracy, going from 81.73\% at the start of test to 90.83\% by the end (shown as the final model $M$'s accuracy in Table \ref{proposed_method}). We observe that $M_a^*+C^*$ does not provide an improvement, but performs similar to $M^*+C^*$ on all performance measures. The numbers reported are calculated as the running average of the different performance measures by evaluating on the fixed size test data (1000 dialogs) once sequentially. We expect an improvement in the various performance measures over longer periods of deployment (test time).

The reward function determines the trade off between the user's task success rate and the human agent's load. We perform additional experiments by modifying the reward function to showcase this trade off. For example, if we want to reduce the load on the human agent further, we can increase the reward provided when the model $M$ is chosen and the model's response is correct and decrease the penalty when the model is chosen and model's response is incorrect. One such reward function is as follows:
\begin{itemize}
    \item +1 : if human $H$ is chosen
    \item +3 : if model $M$ is chosen and the model's response is correct
    \item -3 : if model $M$ is chosen and the model's response is incorrect
\end{itemize}

The results for the new reward function are shown in the last two rows of Table \ref{tab:proposed_method}. In comparison with performance measures for reward function (1,2,-4), for both methods - $M+C^*$ and $M^*+C^*$, we observe a small drop in the user accuracy and a significant increase in model ratio, which showcases our intended goal in altering the reward function.



\section{Related Work}
\label{related_work}
Most of the successful goal-oriented dialog learning systems in the past have been based on slot-filling for domain-specific tasks (\citet{Schatzmann:2006}; \citet{Singh:2000_dialog}). These include Markov Decision Process (MDP) based (\citet{levin_2000}; \citet{Pieraccini:2009}) and Partially Observable Markov Decision Process (POMDP) based (\citet{young2013pomdp}; \citet{W13-4035}) systems. These are Reinforcement Learning (RL) systems that model and track state transitions and take appropriate actions (dialog utterances) to obtain information from the user to fill the desired slots. They require hand crafted features for state and action space representations and hence are restricted to very narrow settings. 

Recently there has been a lot of interest in building end-to-end neural dialog systems for goal-oriented dialog tasks. Both supervised learning based (training the model on collected chat logs of the dialog task) (\citet{bordes2016learning}; \citet{eric2017key}; \citet{wen2016network}) and deep reinforcement learning (RL) based systems (\citet{ZhaoE16}; \citet{LiCLG17}; \citet{PengLLGCLW17}) have been studied. For RL systems, training the model from scratch requires a lot of interactions. Hence, RL systems are often augmented with SL based pre-training on collected chat logs of the dialog task (\citet{Henderson:2008}; \citet{williams2017hybrid};  \citet{lui_lane_2017}).

Training models through RL by using user feedback during deployment makes the system more robust to new user behaviors (\citet{WilliamsZ16}; \citet{lui_lane_2017}). There has also been recent work on actively using human in the loop to teach and assist the learning of neural dialog systems (\citet{LiMCRW16}; \citet{hitlDL}).

While these approaches have focused on different ways to improve the neural goal-oriented dialog systems and maximize user success rate by a) improving the model or, b) better ways of online learning or, c) through human teaching; the problem of handling new user behaviors during deployment has not been solved yet. Our proposed method directly optimizes for maximum user success and provides a framework where existing techniques for model learning, online learning and human teaching can be used in tandem, to enable the end-to-end goal-oriented dialog systems ready for real-world use.


\section{Conclusion and Future Work}
\label{conclusion}
Our proposed method provides a new framework for learning and training goal-oriented dialog systems for the real world. The proposed method allows us to maximize user success rate by minimally using human agents instead of the dialog model for cases where the model might fail. Our evaluation on the modified-bAbI dialog task shows that our proposed method is effective in achieving the desired goals.

We introduce a new method for designing and optimizing goal-oriented dialog systems geared for real-world use. Our method allows the designer to determine the trade-off between the desired user's task success and human agent workload. We believe this opens up a new and promising research direction that would spark an increase in the use of end-to-end goal-oriented dialog systems in the real world soon.

There are several limitations to our current evaluation, which we discuss below and hope to overcome in our future works. While we use simple techniques for the different components in our method, they can be replaced with more sophisticated state of the art techniques for improved performance in terms of absolute values. For example, while we use REINFORCE, an on-policy method for training the classifier, it would be interesting to try off-policy reinforcement learning techniques to use the samples more effectively. We could also try state of the art online learning methods to see how they affect the performance. 

In our experiments, the learning of the classifier $C$ starts from scratch during the deployment. In our future work, we are interested in exploring ways of pre-training the classifier $C$ before deployment, so that the learning of $C$ can happen faster, with less samples during deployment. We are also interested in drawing ideas from novelty detection methods to see if they can help the classifier $C$ to generalize better.

Note that, for our experiments, we use an artificially constructed dataset, modified bAbI dialog tasks, which incorporates two essential assumptions: a) a perfect human-agent and b) correct user feedback. For actual real-world deployments with real users, while the former assumption might still hold true, the latter might not always be true. In our future work, we are interested in relaxing these assumptions and evaluating our proposed method on actual real-world deployments with real users.




\bibliography{tacl2018}
\bibliographystyle{acl_natbib}


\appendix
\section{Appendix: Training Details}
\label{appendix_a}


\begin{table*}[h]
    \centering
    \begin{tabular}{c|c|c|c|c|c}
    \hline
     \textbf{Method} &\multicolumn{2}{c|}{\textbf{User Accuracy}} & \textbf{ Model ratio} & \multicolumn{2}{c}{\textbf{Final Model Accuracy}} \\
     &\textbf{Per-turn} & \textbf{Per-dialog} & & \textbf{Per-turn} & \textbf{Per-dialog}\\ \hline
    Baseline method (M) & 81.73 & 3.7 & 100.0 & 81.73 & 3.7 \\\hline
    \multirow{5}{*}{Reward: 1, 2, -4 ($M^*+C^*$)} & 97.49 & 65.2 & 62.15 & 91.33 & 18.2\\
    &96.4 & 55.1 & 60.45 & 90.58 & 12.7\\
    &94.47 & 38.3 & 71.13 & 89.54 & 9.7\\
    &97.07 & 62.9 & 60.23 & 91.65 & 18.4 \\
    &95.99 & 51.0 & 66.35 & 91.06 & 15.1 \\\hline
    \multirow{5}{*}{Reward: 1, 3, -3 ($M^*+C^*$)} & 92.77 & 29.2 & 73.7 & 88.52 & 11.4\\
    &94.23 & 41.3 & 68.6 & 88.79 & 12.3 \\
    &95.22 & 46.5 & 70.1 & 89.19 & 11.3 \\
    &95.47 & 49.2 & 68.45 & 89.39 & 12.5 \\
    &95.68 & 51.2 & 70.8 & 90.46 & 16.7\\ \hline
    \end{tabular}
    \caption{\textbf{Test results for $M^*+C^*$ method on different permutation of modified-bAbI dialog task's test set}}
    \label{tab:m_star+c_star_all_permutations}
\end{table*}

\begin{table*}[h]
    \centering
    \begin{tabular}{c|c|c|c|c|c}
    \hline
     \textbf{Method} &\multicolumn{2}{c|}{\textbf{User Accuracy}} & \textbf{ Model ratio} & \multicolumn{2}{c}{\textbf{Final Model Accuracy}} \\
     &\textbf{Per-turn} & \textbf{Per-dialog} & & \textbf{Per-turn} & \textbf{Per-dialog}\\ \hline
    Baseline method (M) & 81.73 & 3.7 & 100.0 & 81.73 & 3.7 \\\hline
    \multirow{5}{*}{Reward: 1, 2, -4 ($M_a^*+C^*$)} & 96.15 & 51.8 & 64.64 & 88.52 & 9.9\\
    & 94.69 & 43.4 & 66.96 & 89.16 & 8.8\\
    & 96.52 & 56.5 & 56.57 & 88.75 & 9.6\\
    & 97.99 & 72.9 & 51.26 & 89.37 & 12.5\\
    & 95.63 & 47.6 & 66.27 & 89.12 & 10.5\\ \hline
    \multirow{5}{*}{Reward: 1, 3, -3 ($M_a^*+C^*$)} & 92.51 & 26.9 & 79.16 & 87.81 & 9.9\\
    & 95.23 & 49.3 & 65.67 & 88.8 & 12.0\\
    & 94.03 & 36.4 & 70.06 & 87.82 & 8.2\\
    & 94.55 & 41.4 & 63.03 & 89.73 & 14.8\\
    & 94.1 & 40.0 & 70.57 & 89.57 & 13.2 \\ \hline
    \end{tabular}
    \caption{\textbf{Test results for $M_a^*+C^*$ method on different permutation of modified-bAbI dialog task's test set}}
    \label{tab:adaptive_retrain_batches_all_perm}
\end{table*}

\subsection{Baseline method: ($M$)}
\label{appendix_a1}
\label{sec:baseline-model-hyperparameters}
The hyperparameters used for the training the memN2N model in our baseline method are as follows: hops = 3, embedding\_size = 20, batch\_size = 32. The entire model is trained using stochastic gradient descent (SGD) with learning rate = 0.01 and annealing (anneal\_ratio = 0.5, anneal\_period = 25), by minimizing the standard cross-entropy loss between the predicted response and the correct response. We learn two embedding matrices A and C for encoding context (input and output representations) and a separate embedding matrix B for encoding the query. We use position encoding for encoding word position in the sentence \cite{sukhbaatar2015end}. We also add temporal features to encode information about the speaker for the given utterance (user/system), similar to \cite{bordes2016learning} and weight matrices TA and TC are learned for encoding temporal features. The same weight matrices mentioned above are reused for the 3 hops. We used 599 as the random seed for both tf.set\_random\_seed and tf.random\_normal\_initializer for our embedding matrices. The test results reported for the baseline method are calculated by choosing the model with highest validation per-turn accuracy across multiple runs.

\subsection{Proposed Method: ($M^* + C^*$)}
\label{appendix_a2}
We use the same hyperparameters as the baseline method mentioned above for training the model $M$. The classifier $C$ is trained using REINFORCE \cite{williams1992simple} with a learning rate of 0.01. In $M+C^*$ after every batch of test data, the classifier MLP is updated. In $M^*+C^*$ after every batch of the test data (deployment), along with the classifier MLP update, the model is also updated in the three ways discussed. In $M^*+C^*$ the update using the human responses is done multiple times after every batch (3 in our case). For the update with the training data, two batches of training data are randomly sampled after every batch of test data.

\begin{table}[htp]
    \centering
    \begin{tabular}{c|c|c}
    \hline
     \multicolumn{2}{c|}{\textbf{User Accuracy}} & \textbf{ Model ratio} \\ 
     Per-turn & per-dialog & \\ \hline
    \multicolumn{3}{c}{Baseline method (M)} \\\hline
     81.73 & 3.7 & 100.0 \\\hline
    \multicolumn{3}{c}{Reward: 1, 2, -4, ($M+C^*$)} \\\hline
    92.44 & 33.1 & 50.98 \\
    92.57 & 32.7 & 53.22  \\
    93.96 & 42.0 & 44.63  \\
    94.7 & 43.0 & 45.82  \\
    90.59 & 16.6 & 65.24  \\\hline
    \multicolumn{3}{c}{Reward: 1, 3, -3, ($M+C^*$)} \\\hline
    90.41 & 21.1 & 54.67 \\
    92.25 & 36.1 & 54.79\\
    92.12 & 32.5 & 58.07\\
    89.75 & 18.2 & 65.69\\
    92.05 & 24.6 & 60.92\\\hline
    \end{tabular}
    \caption{\textbf{Test results for $M+C^*$ method on different permutations of modified-bAbI dialog task's test set}}
    \label{tab:m+c_star_all_permutations}
\end{table}

\section{Extended Results}
\label{appendix_b}
Table \ref{tab:m_star+c_star_all_permutations}, \ref{tab:adaptive_retrain_batches_all_perm} and \ref{tab:m+c_star_all_permutations} shows the results for ($M^*+C^*$), ($M_a^*+C^*$) and ($M + C^*$) methods respectively on all the 5 individual permutations of the modified bAbI dialog task test set.

\end{document}